# End-to-end Lung Nodule Detection in Computed Tomography

Dufan Wu[1], Kyungsang Kim[1], Bin Dong[2], Georges El Fakhri[1], and Quanzheng Li[1]

[1] Gordon Center for Medical Imaging, Massachusetts General Hospital and Harvard Medical School, Boston, MA 02114, USA
[2] Beijing International Center for Mathematical Research, Peking University, Beijing 100080, China
`dwu6@mgh.harvard.edu; li.quanzheng@mgh.harvard.edu`

**Abstract.** Computer aided diagnostic (CAD) system is crucial for modern medical imaging. But almost all CAD systems operate on reconstructed images, which were optimized for radiologists. Computer vision can capture features that is subtle to human observers, so it is desirable to design a CAD system operating on the raw data. In this paper, we proposed a deep-neural-network-based detection system for lung nodule detection in computed tomography (CT). A primal-dual-type deep reconstruction network was applied first to convert the raw data to the image space, followed by a 3-dimensional convolutional neural network (3D-CNN) for the nodule detection. For efficient network training, the deep reconstruction network and the CNN detector was trained sequentially first, then followed by one epoch of end-to-end fine tuning. The method was evaluated on the Lung Image Database Consortium image collection (LIDC-IDRI) with simulated forward projections. With 144 multi-slice fanbeam projections, the proposed end-to-end detector could achieve comparable sensitivity with the reference detector, which was trained and applied on the fully-sampled image data. It also demonstrated superior detection performance compared to detectors trained on the reconstructed images. The proposed method is general and could be expanded to most detection tasks in medical imaging.

## 1 Introduction

Computer aided diagnostic (CAD) systems could effectively reduce the intensity of doctors' work by providing fast and high-quality candidates, therefore increase the efficiency of clinical process. Nearly all the current CAD systems operate on the reconstructed images, which are optimized for human observers rather than computers, which could potentially capture details that are subtle to human eyes. In some applications where low-dose or reduced sampling exist, it is often up to the radiologists to resolve the trade-off between accuracy and noise suppressing [1], but the solution may not be the best choice for CAD systems.

In recent years, there were considerably number of works on using deep neural networks as CADs, for various applications including segmentation, detection, diagnosis, etc. [2]. They demonstrated superior performance compared against conventional handcrafted features when given enough training data. However, all these neural networks were still trained from the reconstructed images, thus they may suffer from the non-optimality of the image quality.

Reconstruction algorithms generate images from the original raw data by solving inverse problems. Recently proposed deep-neural-network-based reconstruction algorithms approximate iterative reconstruction process with neural networks, which was trained to minimize the errors between reconstructed images and the ground truth [3][4]. When given enough training data, the deep-neural-network-based reconstruction algorithms could recover more details and maintain better signal-to-noise ratio (SNR) than conventional iterative or image-based methods.

There is an emerging trend on end-to-end signal processing with deep neural networks, and related works include voice recognition, self-driving cars, etc. [5][6] It was demonstrated that end-to-end deep neural networks had improved performance compared to multiple-step learning in these applications.

In this paper, we proposed an end-to-end deep neural network which predicts the location of lung nodules in the computed tomography (CT) images from raw data. The network first converted the raw data to image data with a reconstruction sub-network that approximate a 5-iteration-unrolled primal-dual algorithm; then a 3-dimesional convolutional neural network (3D-CNN) was incorporated as the detection sub-network to locate lung nodules in the images. For more efficient training, the reconstruction sub-network was trained first, followed by the training of the detection sub-network with the reconstructed images. Finally, an end-to-end fine tuning was done on the entire network to maximize the detection performance only. The method was implemented on The Lung Image Database Consortium image collection (LIDC-IDRI) [7], where undersampled CT was simulated from the images with 144 projections per rotation. The proposed method's detection performance and robustness to noises was evaluated and compared against the multiple-step method. The reconstructed images were also analyzed for better understanding of the neural networks.

## 2  Methodology

### 2.1  Overview

Although it is desirable to apply neural networks directly on the raw data, the low coherence between the acquisition and image made local signals in the image domain spreading out in the raw data domain, which lead to difficulty in utilizing the highly efficient CNNs. Furthermore, a detection CAD system should give the position of lesions in the image domain. Hence, our proposed end-to-end network was consisted of the reconstruction sub-network, which mapped the raw data to the image domain; and the detection sub-network, which gave the spatial position of the lesions.

Patch-based detector with fixed window size was used. Denote the raw data as $\mathbf{p}$, the reconstruction sub-network as $R(\mathbf{p}; \boldsymbol{\theta})$, the detection sub-network as $D(\mathbf{x}; \boldsymbol{\eta})$, the patch-wise cross entropy was minimized during training:

$$\boldsymbol{\theta}, \boldsymbol{\eta} = \arg\min \frac{1}{N} \sum_i \sum_j H\big(D(\mathbf{E}_{ij} R(\mathbf{p}_i; \boldsymbol{\theta}); \boldsymbol{\eta}), l_{ij}\big) \quad (1)$$

where $N$ is the total number of patches; $\mathbf{p}_i$ is the raw data of the $i$th scan; $E_{ij}$ is the patch extraction matrix for the $j$th patch from the $i$th scan; $l_{ij}$ is the label of the patch $ij$, which was 1 for patches containing nodule centers and 0 for the rest. $H(\cdot,\cdot)$ is the cross entropy loss.

Directly solving (1) yields slow training due to the heavy computational loads of the reconstruction sub-network and the relatively slow convergence of the detection sub-network. To improve training speed, the training was split into 3 steps: training of the reconstruction sub-network, training of the detection sub-network, and end-to-end fine tuning.

### 2.2 Reconstruction Sub-Network

The reconstruction sub-network $R$ was firstly trained to minimize the L2 error between the reconstructed images and the ground truth:

$$\boldsymbol{\theta}_1 = \arg\min \frac{1}{N_1} \sum_i \|R(\mathbf{p}_i; \boldsymbol{\theta}) - \mathbf{x}_{Ti}\|_2^2 \quad (2)$$

where $N_1$ is the total number of training images and $\mathbf{x}_{Ti}$ is the ground truth image corresponding to the $i$th scan.

Many choices of $R$ exist, most of which realize finite iteration of existing algorithms with neural network, by replacing the part related to the prior term with trainable CNNs [3][4]. The entire neural network could then be trained through (2). Thorough studies are yet to be done for comparison between various network structures, and in this study the primal-dual framework was incorporated [3]. 5 unrolled iterations were used, and the model parameters were chosen so that the training could be accomplished in a reasonable time.

Furthermore, due to the relative large memory footprint of the reconstruction sub-network, it was infeasible to feed it with large number of slices. Instead, the primal-dual network took only a few adjacent slices (e.g. 3), and the final reconstructed images was the aggregation from different slices. Denote the primal-dual network as $R_{PD}(\mathbf{p}; \boldsymbol{\theta})$, then $R(\mathbf{p}; \boldsymbol{\theta})$ could be written as:

$$R(\mathbf{p}; \boldsymbol{\theta}) = \frac{\sum_k \mathbf{W}_k^T R_{PD}(\mathbf{W}_k \mathbf{p}; \boldsymbol{\theta})}{\sum_k \mathbf{W}_k^T \mathbf{W}_k \mathbf{1}} \quad (3)$$

where $\mathbf{W}_k$ is the matrix to extract the $k$th sub-slices from the raw data $\mathbf{p}$. $\mathbf{1}$ is an all-ones matrix with the same size of $\mathbf{p}$. The structure of the primal-dual network is demonstrated in figure 1.

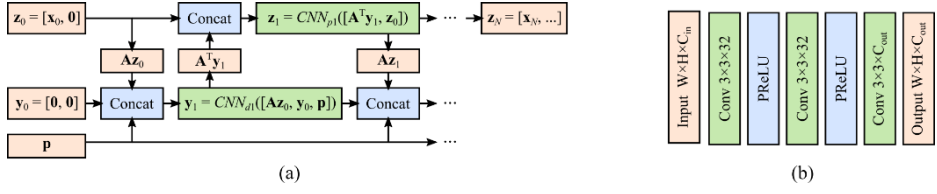

**Fig. 1.** The structure of the primal-dual network: (a) general structure; (b) structure of $CNN_{pi}$ and $CNN_{di}$. $z_i$ and $y_i$ are primal and dual variables respectively, and we used $N_{primal} = N_{dual} = 2$ in the study. $x_0$ had 3 slices (channels), so $z_i$ and $y_i$ both had 6 channels. $x_N$ is the reconstructed image. $C_{out}$ equals to the number of channels of $z_i$ or $y_i$.

### 2.3 Detection Sub-Network

After training of $R(p; \theta_1)$, patches were randomly extracted from the reconstructed images and the detection sub-network $D$ was trained as:

$$\eta_1 = \arg\min \frac{1}{N} \sum_i \sum_j H\big(D(E_{ij} R(p_i; \theta_1); \eta), l_{ij}\big) \tag{4}$$

where the notations were the same with that of (1), except that only $\eta$ was optimized.

The detector incorporated in this work was the 3D-CNN proposed in [8]. A patch was considered positive if the center of a non-small lung nodule is within the patch. Flips along the three axes were used for data augmentation because of its simplicity, but any augmentation that is expressible by matrix multiplication could fit into the proposed framework. The detection network is demonstrated in figure 2.

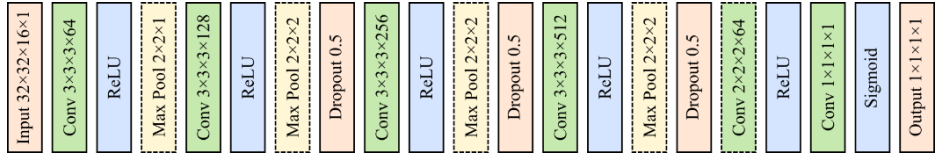

**Fig. 2.** The structure of the detection network. The modules with the dashed boxes did not use padding, and the rest used zero padding if applicable.

### 2.4 End-to-End Fine Tuning

After the two sub-networks were sequentially trained, one epoch of training of (1) was carried out with initial values set as $\theta_1$ and $\eta_1$. The gradient backpropagation from the detection sub-network to the reconstruction sub-network could be derived by chain rule of derivatives.

### 2.5 Inference

During inference from raw data $p$, the detector $D$ was applied on sliding windows on $R(p; \theta)$, followed by a non-max suppressing (NMS) step to get the final detection.

## 3 Simulation Setup

### 3.1 Data Source

The LIDC-IDRI dataset was used for the simulation. It contains 1,018 chest CT scans from various scanners and each image was annotated by 4 radiologists. The detection task was set to detect the non-small nodules (nodules with ≥ 3mm diameter), because imaging of small nodules is not stable for the used CT protocols.

All the images were resampled to $1 \times 1 \times 2$ mm$^3$, and forward projected in a multi-slice fanbeam geometry with equal angular detectors. 144 projections per rotation were used, and the detector had 736 units per layer with a pixel size of $1.2858 \times 2$ mm$^2$. The source-center and source-detector distances were 595 mm and 1086.5 mm.

The neural networks were trained on noiseless simulated data, but Poisson noise with equivalent initial photon number of $1 \times 10^5$ and $5 \times 10^4$ per ray were added at the test time to evaluate the robustness of the neural networks against inconsistency.

### 3.2 Training Parameters

The dataset was split into the training set and testing set with 916 and 102 scans respectively. The training parameters for the neural networks were as follows:

**Reconstruction Sub-Network.** The primal-dual network $R_{PD}$ took 3 adjacent layers as input and realized 5 iterations of the primal-dual algorithm. The initial images were taken as the filtered backprojection (FBP) results. Adam optimizer with learning rate of $1 \times 10^{-4}$ was used, with $\beta_1 = 0.9$ and $\beta_2 = 0.999$. 50 samples were randomly extracted from each scan, and 1 epoch was run for the training.

**Detection Sub-Network.** A patch size of $32 \times 32 \times 16$ was used for sampling. For each annotation on non-small nodules, the sampling was augmented by randomly translation between [-8, 8] mm and flipping along 3 axes for 20 times, which generated 60 to 80 positive samples for each nodule (each nodule was annotated 3 to 4 times by different radiologists).

All the negative samplings kept a safe margin of 64 mm from any positive samplings. A 5-time augmented sampling was done for each non-nodule annotation. 400 patches were randomly extracted inside the lung whereas 100 patches were randomly extracted on the edge of the lung mask for each scan. The mask was derived from the FBP results.

The same Adam optimizer in the reconstruction sub-network was used here. 10 epochs of training were done with a minibatch size of 50.

**End-to-End Fine Tuning.** The same patch sampling coordinates were used with that for the detection sub-network. For each minibatch, 32 layers were extracted, and Adam optimizer was applied to the patches within the extracted layers. For each scan, multiple sub-layer extractions were done to cover all the patches once. 1 epoch of fine tuning was done with the same Adam optimizer in the previous steps.

### 3.3 Evaluation

For each testing scans, the detector was applied on sliding windows inside the lung masks with step size of 4 mm. NMS was then applied with intersection over union (IoU) threshold of 0.5. Free-response receiver operator curve (FROC) analysis was done with 1,000 bootstrapping to evaluate the detection performance [9], where a true positive was count if a nodule center was within the positive patch. Mean FROC scores were also calculated as the mean value of the sensitivities at 1/8, 1/4, 1/2, 1, 2, 4, 8 false positives per scan.

## 4 Results

### 4.1 FROC analysis

Figure 3 gave the results of FROC for various detectors under 3 different noise levels. The end-to-end method was compared against the detector based on the FBP results and the two-step approach, where the primal-dual network was trained first and the detector was trained on its reconstruction results. Furthermore, the same detector was trained on the original resampled images to provide a reference FROC performance.

The noiseless FROC results demonstrated significant improvement of the end-to-end approach over the two-step approach. Results from the proposed method was also comparable to the reference results when average false positives per scan were within [1,4]. When small amount of noise ($N_0 = 1 \times 10^5$) was added, the performance of end-to-end detector remained almost the same, whereas the separately trained detectors' performance was obviously decreased. The performance of the end-to-end detector further deteriorated when the noise level increased, but its advantage over the separately trained detector was maintained.

The mean FROC scores are listed in table 1. It could be noted that end-to-end detector had a higher FROC score when $N_0 = 1 \times 10^5$ compared to the noiseless situation. It indicated that the noise had little influence on the detector, and the gain of score was due to normal detector performance noises.

Table 1. Mean FROC scores

| Noise level | Reference | FBP | Two-step | End-to-end |
| --- | --- | --- | --- | --- |
| None | 0.636 | 0.563 | 0.560 | 0.608 |
| $N_0 = 1 \times 10^5$ | N/A | 0.549 | 0.538 | 0.615 |
| $N_0 = 5 \times 10^4$ | N/A | 0.525 | 0.512 | 0.587 |

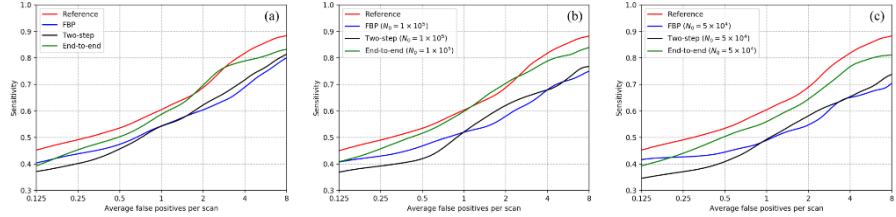

**Fig. 3.** FROC of different detectors: (a) noiseless performance; (b) performance with $1 \times 10^5$ photons per ray; (c) performance with $5 \times 10^4$ photons per ray. "Reference" refers to detection on original resampled images; "FBP" refers to detection on FBP results; "Two-step" refers to detection on primal-dual results; "End-to-end" refers to the proposed method. Only the mean value from the bootstrapping was shown for better visual effect.

### 4.2 Reconstructed Images

One slice from the reconstructed images is shown in figure 4, where the contrast to noise ratio (CNR) was calculated for a nodule in the slice. The two-step result had the best visual performance, whereas the end-to-end result had the undersampling streak artifacts. However, the better detection performance of the end-to-end detector indicated the difference between human observer and the computer vision, where the latter could ignore such structured noise for the detection task.

Though there existed subtle visual differences between the two-step and end-to-end results except for the streak artifacts, the gain on CNR was significant for the end-to-end result, which could be one of the reasons lead to better detection performance. Note that both two-step and end-to-end results had larger CNR than the reference result. The former was because of the existence of noises in the original images, the latter was because of the gain in contrast.

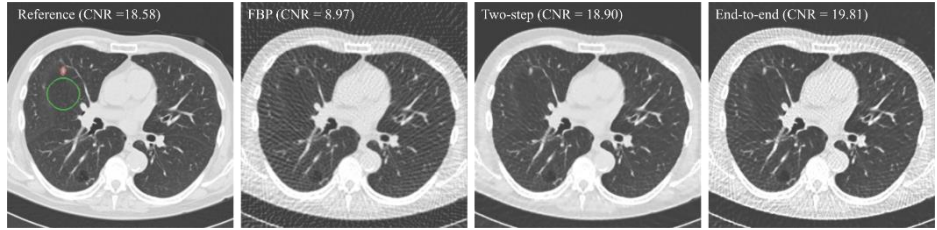

**Fig. 4.** Reconstructed axial images from different methods. The CNRs were calculated from the nodule within the red circle against the background within the green circle. The display window is [-1400, 200] HU.

## 5 Conclusion and Discussion

In this paper we proposed an end-to-end lung nodule detection system for undersampled CT data, where a reconstruction sub-network and a detection sub-network were trained end-to-end to maximize the detection performance. The end-to-end methods achieved comparable performance with the detector trained on the fully-sampled data and had a significant gain over two-step methods. The proposed method is general and could be easily applied to other detection tasks and other modalities in medical imaging.

Although the training relied on simulation data, its robustness against moderate noise was also validated in the simulation. The results indicated that if the simulation does not significantly bias from the reality, the trained neural network should be generalizable enough to assure acceptable performance on real data.

Training-from-scratch had the potential to reach better solution than the current fine-tuning scheme, but it was too time consuming because of the heavy computational load of the training of reconstruction network, which required the entire slice as input and could not be broken into patches.

It is acknowledged that neither the reconstruction nor the detection sub-networks were carefully chosen or optimized in the current study. But the advantage of the end-to-end method should be maintained with different reconstruction or detection modules, since the solution to the two-step method could be further optimized in the end-to-end framework. We are actively working with more advanced detection and reconstruction neural networks [10].